\documentclass[10pt,twocolumn,letterpaper]{article}

\usepackage{cvpr}              

\usepackage{graphicx}
\usepackage{amsmath}
\usepackage{amssymb}
\usepackage{booktabs}
\usepackage{flushend}

\usepackage[pagebackref,breaklinks,colorlinks]{hyperref}

\usepackage[capitalize]{cleveref}
\crefname{section}{Sec.}{Secs.}
\Crefname{section}{Section}{Sections}
\Crefname{table}{Table}{Tables}
\crefname{table}{Tab.}{Tabs.}

\def\cvprPaperID{*****} 
\def\confName{CVPR}
\def\confYear{2022}

\providecommand{\keywords}[1]{\textbf{\textit{Index terms ---}} #1}

\begin{document}

\title{SFU-HW-Tracks-v1: Object Tracking Dataset on Raw Video Sequences}

\author{Takehiro Tanaka\\
School of Engineering Science\\
Simon Fraser University\\
{\tt\small takehiro\_tanaka@sfu.ca}
\and
Hyomin Choi\\
InterDigital and\\
Simon Fraser University\\
{\tt\small chyomin@sfu.ca}
\and
Ivan V. Baji\'{c}\\ 
School of Engineering Science\\
Simon Fraser University\\
{\tt\small ibajic@ensc.sfu.ca}
}
\maketitle

\begin{abstract}
   We present a dataset that contains object annotations with unique object identities (IDs) for the High Efficiency Video Coding (HEVC) v1 Common Test Conditions (CTC) sequences. Ground-truth annotations for 13 sequences were prepared and released as the dataset called SFU-HW-Tracks-v1. For each video frame, ground truth annotations include object class ID, object ID, and bounding box location and its dimensions. The dataset can be used to evaluate object tracking performance on uncompressed video sequences and study the relationship between video compression and object tracking.
\end{abstract}


\section{Introduction}
\label{sec:intro}

There is an increasing interest in the interplay between image/video compression and computer vision~\cite{khatoonabadi2013video,choi2017corner,choi2017hevc,alvar2018cantell,alvar2018canfind,torfason2018towards,choi2018high,MV-YOLO,dfc_icip_2018,saeed_multi_task_learning,LSS_ICIP_2021,choi_scalable_2021}. In order to study their relationship, one must have task-relevant labels on raw (uncompressed) images and video. Currently, there is a lack of such datasets, especially for video.

Our group has previously released a dataset called SFU-HW-Objects-v1~\cite{SFU-HW-Objects-v1}, which is currently being used in MPEG Video Coding for Machines (VCM) standardization~\cite{VCM-EE-m58474}. This dataset  provides object class labels and bounding boxes for objects found in the High Efficiency Video Coding (HEVC) Common Test Conditions (CTC) video sequences~\cite{hevc_ctc}. This made it possible for object detection models to be trained and tested on HEVC CTC video sequences using labels provided in SFU-HW-Objects-v1. However, it is not possible to train or test object tracking models on annotations in SFU-HW-Objects-v1, because there are no object identifiers in in that dataset. For example, if a video frame shows three people, the annotations in SFU-HW-Objects-v1 would have three bounding boxes, each with a class label “Person”, but there is no annotation to distinguish these three persons from each other. In the next frame, one would also have three bounding boxes, each associated with label “Person” and no information which person is which, so it is impossible to track a specific person from frame to frame. 

The new dataset, \textbf{SFU-HW-Tracks-v1}, extends the previous dataset (SFU-HW-Objects-v1) by providing a unique object identifier for each object in the dataset. In the above example, this would mean that the three people are labeled ``Person(0),'' ``Person(1),'' and ``Person(2),'' and these persons have the same labels in the next frame, so it is possible to track how each person moves from frame to frame. This enables tracking models to be trained and tested on SFU-HW-Tracks-v1, which was not possible on the previous dataset. SFU-HW-Tracks-v1 is available at \url{https://doi.org/10.17632/d5cc83ks6c.1}

\section{Data description}
\label{sec:data_description}
We prepared object tracking annotations for 13 HEVC CTC video sequences~\cite{hevc_ctc}, as shown in Table~\ref{tab:sequences}. These sequences are uncompressed, in the YUV420 format, and can be acquired from Joint Collaborative Team on Video Coding (JCT-VC). For each video frame, ground truth annotations include object class ID, object ID, and bounding box location and its dimensions. 

\begin{table*}[t]
  \centering
  \begin{tabular}{@{}cccccclc@{}}
    \toprule
    Class & Sequence & Resolution & Frame count & Frame rate (Hz) & Bit depth & Class IDs & \# Object classes\\
    \midrule
    B & BasketballDrive & 1920$\times$1080 & 500 & 50 & 8 & \{0, 32, 56\} & 4\\
    B & Cactus & 1920$\times$1080 & 500 & 50 & 8 & \{56\} & 1\\
    B & Kimono & 1920$\times$1080 & 240 & 24 & 8 & \{0, 26\} & 2\\
    B & ParkScene & 1920$\times$1080 & 240 & 24 & 8 & \{0, 1, 13\} & 4\\
    C & BasketBallDrill & 832$\times$480 & 500 & 50 & 8 & \{0, 32, 56\} & 4\\
    C & PartyScene & 832$\times$480 & 500 & 50 & 8 & \{0, 41, 58, 74, 77\} & 6\\
    C & RaceHorsesC & 832$\times$480 & 300 & 30 & 8 & \{0, 17\} & 2\\
    D & BasketBallPass & 416$\times$240 & 500 & 50 & 8 & \{0, 32, 56\} & 4\\
    D & BlowingBubbles & 416$\times$240 & 500 & 50 & 8 & \{0, 41, 77\} & 3\\
    D & RaceHorsesD & 416$\times$240 & 300 & 30 & 8 & \{0, 17\} & 2\\
    E & KristenAndSara & 1280$\times$720 & 600 & 60 & 8 & \{0, 63, 67\} & 3\\
    E & Johnny & 1280$\times$720 & 600 & 60 & 8 & \{0, 27, 63\} & 3\\
    E & FourPeople & 1280$\times$720 & 600 & 30 & 8 & \{0, 41, 56, 58\} & 4\\
    \bottomrule
  \end{tabular}
  \caption{List of HEVC CTC sequences for which object tracking ground truth is provided.}
  \label{tab:sequences}
\end{table*}

\begin{table}[t]
  \centering
  \begin{tabular}{@{}clcl@{}}
    \toprule
    Class ID & Object & Class ID & Object\\
    \midrule
    0 & Person & 41 & Cup\\
    1 & Bicycle & 56 & Chair\\
    13 & Bench & 58 & Potted plant\\
    17 & Horse & 63 & Laptop\\
    26 & Handbag & 67 & Cell phone\\
    27 & Tie & 74 & Clock\\
    32 & Sports ball & 77 & Teddy bear\\
    \bottomrule
  \end{tabular}
  \caption{List of object class IDs in the ground truth.}
  \label{tab:class_IDs}
\end{table}

\begin{figure*}
    \centering
    \includegraphics[width=0.99\textwidth]{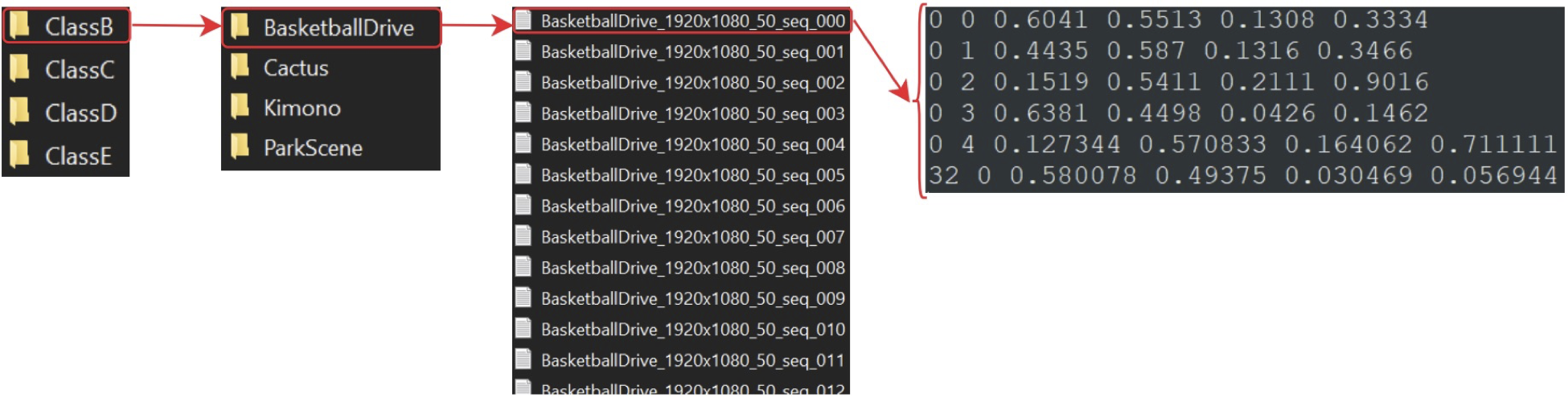}
    \caption{Folder and file structure.}
    \label{fig:folder_structure}
\end{figure*}

The dataset has separate folders for each sequence class (B, C, D, E), which differ in resolution, and each class folder contains individual sequence folders, as shown in Figure~\ref{fig:folder_structure}. Each sequence folder contains one annotation file per frame, which is a text file and can be viewed in any text editor. Each row in the annotation file corresponds to an object in the corresponding frame, and contains the following information: 
$$[\text{Class ID}, \text{ Object ID}, \; x, \; y, \; w, \; h].$$

Class ID represents the identifier of an object class, for example “Person,” “Bicycle,” etc. All the class IDs in the ground truth are listed in Table~\ref{tab:class_IDs}, and they are all part of Common Objects in Context (COCO)~\cite{COCO} object classes. Object ID refers to the unique identity of each object. For example, if a frame contains two persons, unique IDs are provided for each person, so they can be distinguished as ``Person(0)'' and ``Person(1).'' Finally, $x$ and $y$ are the horizontal and vertical coordinates of the object’s bounding box in relative coordinates (relative to the frame dimensions, as explained below), while $w$ and $h$ are the relative dimensions of the bounding box. The center position of the object's bounding box in relative coordinates is obtained from the absolute coordinates $x^*$ and $y^*$ (from the top-left corner), and frame width $N$ and height $M$, as:
\begin{equation}
    x = \frac{x^*}{N}, \qquad y = \frac{y^*}{M}.
\end{equation}
Similarly, relative bounding box width and height, $w$ and $h$, are obtained from the absolute width and height, $w^*$ and $h^*$, as:
\begin{equation}
    w = \frac{w^*}{N}, \qquad h = \frac{h^*}{M}.
\end{equation}

\begin{figure*}
    \centering
    \includegraphics[width=0.99\textwidth]{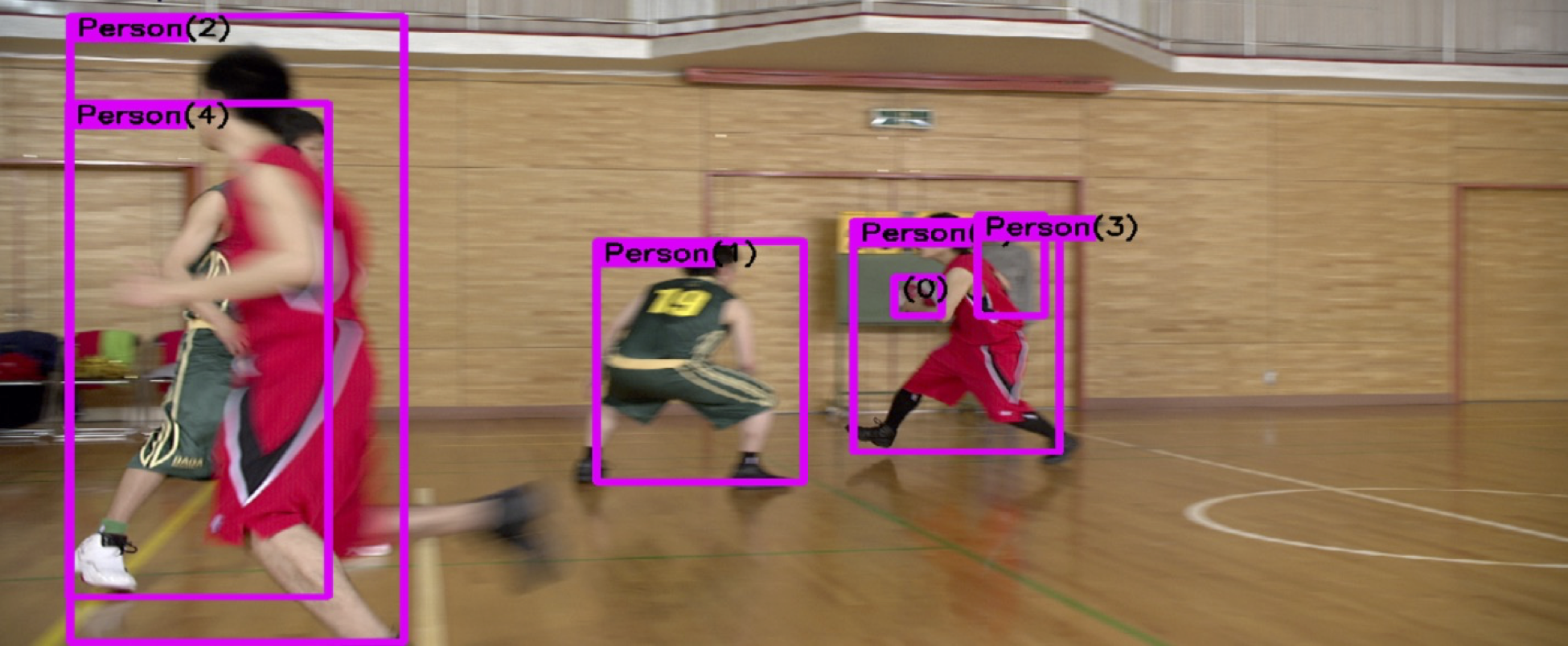}
    \caption{Sample frame from the sequence BasketballDrive with overlaid ground-truth annotations.}
    \label{fig:gt_example}
\end{figure*}

The corresponding annotations can be visualized overlaid on the image frame using \texttt{Yolo\_mark}\footnote{We slightly modified \texttt{Yolo\_mark} to show the Object ID for each object.}~\cite{Yolo_mark}, as shown in Figure~\ref{fig:gt_example}. The figure shows a frame from the BasketballDrive sequence, which has four objects from the ``Person'' class (Class ID 0), with Object IDs from 0 to 3, so they appear as ``Person(0)'' to ``Person(3).'' There is also a single ``Sports ball'' object (Class ID 32) with Object ID 0. The combination of Class ID and Object ID uniquely identifies each annotated object.

\begin{figure*}
    \centering
    \includegraphics[width=0.99\textwidth]{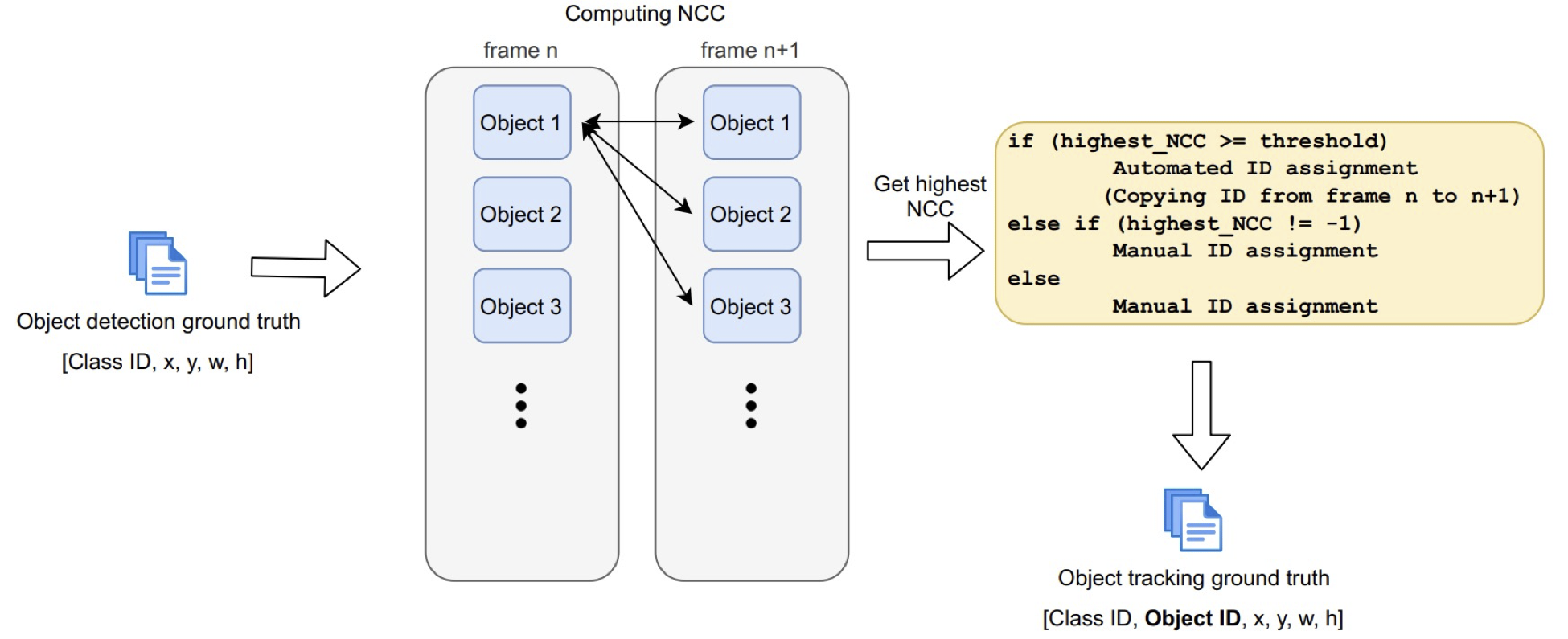}
    \caption{Semi-automated annotation process for assigning Object IDs.}
    \label{fig:annotation_process}
\end{figure*}

\section{Experimental design, materials and methods}
\label{sec:methods}

Tracking annotations in SFU-HW-Tracks-v1 were created based on object detection annotations in SFU-HW-Objects-v1~\cite{SFU-HW-Objects-v1}, which contain the following information for each object: 
$$[\text{Class ID}, \; x, \; y, \; w, \; h].$$
However, SFU-HW-Objects-v1 is not suitable for tracking purposes because there is no annotation distinguishing different objects from the same class. Therefore, we further created unique Object IDs within each class, which enables distinction of different objects in each class. Further, the same Object ID is used for the same object in different frames, which allows computing tracking metrics. These object IDs are included in the second column of the provided annotation files. 

We used normalized cross-correlation (NCC)~\cite{NCC_ICASSP_2006} to measure the similarity between two bounding boxes, where each contains an object. To find matching locations for objects in neighboring frames ($n$ and $n+1$), we computed NCC for all possible pairs of object bounding boxes between these two frames. For each object bounding box in frame $n$, we took as its best match the box in frame $n+1$ that gave the highest NCC score. If the NCC score was greater than the threshold value ($0.6$ in most sequences), we copied the corresponding Object ID from frame $n$ to the best-matched box in frame $n+1$. If the NCC score of the best-matched box was less than the threshold value, we manually assigned an Object ID to that object in frame $n+1$ after visual inspection. The threshold value was manually adjusted in the range $[0.55, 0.75]$ in several sequences to account for different characteristics of objects and their appearance. 

If a particular object did not exist in frame $n$ but was found in frame $n+1$ (e.g., the object has entered the scene), we defined the corresponding NCC score as $-1$. In this case, we manually assigned an Object ID for the object in frame $n+1$. Such situation could occur when an object disappears and re-appears due to occlusion, or appears for the first time. The manual Object ID assignment was conducted after visualizing the annotations on the frame using \texttt{Yolo\_mark}, comparing the bounding boxes, and/or using the object annotation files. After assigning the Object IDs in the current frame, the annotation process proceeded to the next frame. Figure~\ref{fig:annotation_process} summarizes the semi-automated process of assigning Object IDs.

\section{Ethics statement}

No human, animal subjects, and data from social media platforms were involved in this work.

\section{CRediT author statement}

\textbf{Takehiro Tanaka}: Methodology, Software, Data curation, Writing - Original draft, Visualization. \textbf{Hyomin Choi}: Conceptualization, Methodology, Software, Data curation, Supervision, Writing - Review \& Editing. \textbf{Ivan V. Baji\'{c}}: Conceptualization, Supervision, Writing - Review \& Editing, Project administration, Funding acquisition.

\section{Acknowledgment}
The authors would like to thank Timothy Woinoski of Simon Fraser University for his help on tracking annotations. 

The funding for this work was provided by the Natural Sciences and Engineering Research Council (NSERC) of Canada, under the grant RGPIN-2021-02485.

{\small
\bibliographystyle{ieee_fullname}
\bibliography{egbib}
}

\end{document}